\documentclass[11pt]{article}

\usepackage[utf8]{inputenc}
\usepackage[T1]{fontenc}
\usepackage{hyperref}
\usepackage{url}
\usepackage{booktabs}
\usepackage{amsfonts}
\usepackage{nicefrac}
\usepackage{microtype}
\usepackage{amsmath, amssymb}
\usepackage{graphicx}
\usepackage{geometry}
\usepackage{authblk}
\usepackage{xcolor}
\usepackage{subcaption}

\usepackage{listings}
\usepackage{xcolor}

\lstdefinestyle{mycode}{
    language=Python,
    basicstyle=\ttfamily\small, 
    keywordstyle=\color{blue}\bfseries,
    commentstyle=\color{gray}\itshape,
    stringstyle=\color{purple},
    breaklines=true,            
    showstringspaces=false,     
    numbers=none                
}

\graphicspath{{./fig_src/}}

\title{Phasor Attention:\\ Mean Root Square Normalization for Phase Manifold Preservation}

\author[1]{\textbf{Sungwoo Goo}}
\author[1]{\textbf{Hwi-yeol Yun}}
\author[2]{\textbf{Sangkeun Jung}}

\affil[1]{College of Pharmacy, Chungnam National University, Daejeon, Republic of Korea}
\affil[2]{Department of Computer Science \& Engineering, Chungnam National University, Daejeon, Republic of Korea}

\affil[ ]{\small \texttt{swgoo91@gmail.com, hyyun@cnu.ac.kr, hugmanskj@gmail.com}}
\date{}

\begin{document}

\maketitle

\begin{abstract}
While Root Mean Square Normalization has become the \textit{de facto} standard for accelerating modern sequence models, its reliance on the quadratic accumulation of independent scalars ($\sum x^2$) inherently triggers outlier-induced numerical instability, gradient starvation, and anisotropic phase distortion.
We introduce \textbf{Mean Root Square Normalization (MRSNorm)}.
By structurally pairing channels into 2D phasors, MRSNorm mathematically inverts the traditional scaling paradigm: it computes the localized $L_2$ magnitudes (Root Square) before aggregating them via a global $L_1$ average (Mean). 

This operational inversion strictly constrains activations to a phasor manifold, preserving conformal invariance. By sharing a single affine weight across phasor components, MRSNorm \textbf{halves the total number of learnable parameters}, proving that unconstrained spatial scaling in standard norms is a harmful redundancy.
We analytically demonstrate that this geometric constraint yields a built-in, trigonometric gradient clipper governed by the Pythagorean identity, unconditionally equalizing the local gradient norm to ensure \textit{Gradient Homogeneity}. 

Empirical evaluations on a ResNet with CIFAR-100 show that despite halved parameters, MRSNorm provides critical structural stability under rigorous stress tests.
Under extreme hyperparameter settings where standard normalizations suffer from gradient divergence, MRSNorm successfully prevents numerical explosion and secures stable optimization trajectories.
Our findings propose a fundamental paradigm shift toward \textit{phasor-based deep representation learning}.
The implementation of MRSNorm is available at Appendix~\ref{sec:imp}.
\end{abstract}

\section{Introduction}
Normalization layers are the indispensable backbone of modern deep learning, essential for mitigating internal covariate shift and stabilizing the training of exceedingly deep networks.
While Root Mean Square Normalization (RMSNorm) \cite{zhang2019root} has been widely adopted in Large Language Models (LLMs) for its computational efficiency over Layer Normalization (LN) \cite{ba2016layer}, it has a critical structural vulnerability: \textbf{the quadratic accumulation of variances} ($\sum x^2$).
In mixed-precision environments, a single magnitude outlier can disproportionately dominate the denominator, triggering catastrophic numerical explosions and inducing gradient starvation for remaining channels.

To address these limitations, we propose \textbf{Mean Root Square Normalization (MRSNorm)}. 
MRSNorm mathematically inverts the order of operations.
By pairing adjacent channels into $2D$ phasors, MRSNorm first computes localized $L_2$ magnitudes (Root Square) and subsequently aggregates them via a global $L_1$ average (Mean). 

MRSNorm is also parameter-efficient by sharing a single affine weight ($\gamma$) across $2D$ phasor components, it \textbf{halves the total learnable parameters}, eliminating the redundant degrees of freedom that distort phase.
Crucially, as derived in Section~\ref{sec:grad}, the Pythagorean identity ($\cos^2\theta + \sin^2\theta \equiv 1$) ensures that the local gradient update magnitude for each phasor is equalized.
This structurally guaranteed \textbf{Gradient Homogeneity} functions as a built-in gradient clipper, preventing starvation without distorting the directional optimization trajectory.

\section{Related Works}
The quest for stable, efficient deep learning models has driven the continuous evolution of normalization techniques and normalization-free architectures.
Our proposed MRSNorm stands at the intersection of geometric phase-coding and robust gradient stabilization, addressing the limitations of prior methodologies.

\subsection{Evolution of Normalization Layers}
Layer Normalization (LN) \cite{ba2016layer} has been the cornerstone of Transformer architectures, ensuring stable forward dynamics by normalizing activations across the channel dimension with mean-centering and variance scaling.
However, the computational overhead of calculating the mean led to the development of RMSNorm \cite{zhang2019root}.
RMSNorm demonstrated that the mean-centering operation is largely dispensable for success, effectively accelerating training throughput.
Consequently, RMSNorm has become the standard for modern Large Language Models (LLMs). 

Despite its computational advantages, RMSNorm introduces a structural vulnerability: the unbounded quadratic accumulation of channel magnitudes ($\sum x^2$).
This mechanism makes the denominator highly susceptible to activation outliers, resulting in frequent numerical explosions (NaNs) and gradient starvation for low-magnitude channels \cite{sun2024massive}.
MRSNorm resolves this by inverting the operation sequence to a linear mean of localized $L_2$ phasors, eliminating the risk of catastrophic variance explosion while retaining the zero-centering computational efficiency of RMSNorm.

\subsection{Phase and Complex-Valued Representations}
The encoding of information using phase and complex numbers has a history in signal processing and deep learning.
Complex-valued Neural Networks \cite{trabelsi2017deep} attempted to leverage phase information by substituting real numbers with complex arithmetic.
However, without stringent geometric constraints on the amplitude manifold, these networks effectively defaulted to 2-channel real-valued operations, failing to mitigate the vanishing and exploding gradient problems. 

A notable attempt to preserve directional information was the \textbf{Squashing function} introduced in Capsule Networks \cite{sabour2017dynamic}, which maps the vector length to a range between 0 and 1 while preserving its orientation.
However, as we theoretically analyzed in Section~\ref{sec:amp_block}, instance-wise amplitude normalization (e.g., Squashing or direct unit-norm projection) inherently suffers from a \textbf{Radial Gradient Blockade}, where the Jacobian becomes singular in the direction of the vector's magnitude, thereby stifling learning signals in deep architectures.

More recently, RoPE \cite{su2024roformer} successfully introduced phase-domain rotations to encode relative positional information without altering the vector norm. but this approach is limited to positional encoding and does not address the broader issues of gradient starvation and numerical instability in deep networks.

Unlike individual-based squashing or projection methods, MRSNorm employs an \textbf{ensemble-level homeostatic scaling}, preserving the essential radial gradient pathways while enforcing an \textbf{gradient homogeneity} across network depths.

\subsection{Gradient Clipping and Optimization Stability}
Gradient clipping \cite{pascanu2013difficulty} remains a ubiquitous heuristic to prevent gradient explosion in deep and recurrent architectures.
However, traditional magnitude-based clipping blindly truncates the gradient vectors, distorting the critical directional and angular information essential for accurate geometric optimization \cite{chen2020understanding}.
Alternatively, adaptive gradient scaling methods (e.g., AdamW \cite{loshchilov2017decoupled}) attempt to smooth the optimization trajectory but frequently suffer from a generalization gap compared to Stochastic Gradient Descent (SGD) \cite{wilson2017marginal}, as they tend to warp the update directions of multidimensional features anisotropically.

MRSNorm introduces a fundamentally different paradigm to address this optimization instability.
Rather than relying on heuristic, we mathematically enforce \textbf{Gradient Homogeneity} at the microscopic phasor level.
As we formally derive in Section~\ref{sec:grad}, our $2D$ phasor grouping guarantees that the local gradient update magnitude of each individual phasor is unconditionally equalized, rendering the gradient scale invariant to local coordinate magnitudes.
This structural guarantee ensures the gradient flow across all channels, preventing gradient starvation and paving the way for smooth convergence into flat, robust minima under deeply-layered conditions.

\section{Methods}
Given an input feature vector $\mathbf{h} \in \mathbb{R}^N$ (where $N$ is even), we restructure the feature space into $P = N/2$ pairs of 2D phasors. Let $\mathbf{z}_i = [x_i, y_i]^T \in \mathbb{R}^2$ denote the $i$-th phasor pair for $i \in \{1, \dots, P\}$. 

MRSNorm deliberately inverts the traditional scaling paradigm.
Rather than calculating a global quadratic mean of individual scalars, we first compute the localized $L_2$ magnitude (Root Square) for each phasor, followed by a global $L_1$ average (Mean) across all $P$ pairs:
\begin{equation}
    \label{eq:mrsnorm}
    S_{MRS} = \frac{1}{P} \sum_{i=1}^P \| \mathbf{z}_i \|_2 + \epsilon = \frac{1}{P} \sum_{i=1}^P \sqrt{x_i^2 + y_i^2} + \epsilon
\end{equation}
where $\epsilon$ is a small constant added for numerical stability.

The normalized phasor outputs $\mathbf{\hat{z}}_i = [\hat{x}_i, \hat{y}_i]^T$ are then obtained via isotropic vector scaling:
\begin{equation}
    \mathbf{\hat{z}}_i = \gamma_i \frac{\mathbf{z}_i}{S_{MRS}}
\end{equation}
where $\gamma_i \in \mathbb{R}$ is a affine parameter applied uniformly to the $i$-th phasor pair. 

By enforcing a unified scale factor ($S_{MRS}$) and sharing the affine weight ($\gamma_i$) across the $2D$ plane, MRSNorm strictly preserves the exact phase angle ($\theta_i$) of the original phasor: 
\begin{equation}
    \theta(\mathbf{\hat{z}}_i) = \arctan\left(\frac{\hat{y}_i}{\hat{x}_i}\right) = \arctan\left(\frac{y_i}{x_i}\right) = \theta_i
\end{equation}

This design choice not only maintains conformal invariance but also intrinsically halves the total number of affine parameters ($N/2$) compared to standard normalizations.
Consequently, MRSNorm eradicates the harmful degrees of freedom that traditionally distort geometric phase information, proving that unconstrained spatial scaling is a redundant source of instability.

\section{Experiments}
\label{sec:experiments}

\begin{figure}
    \centering
    \includegraphics[width=.98\textwidth]{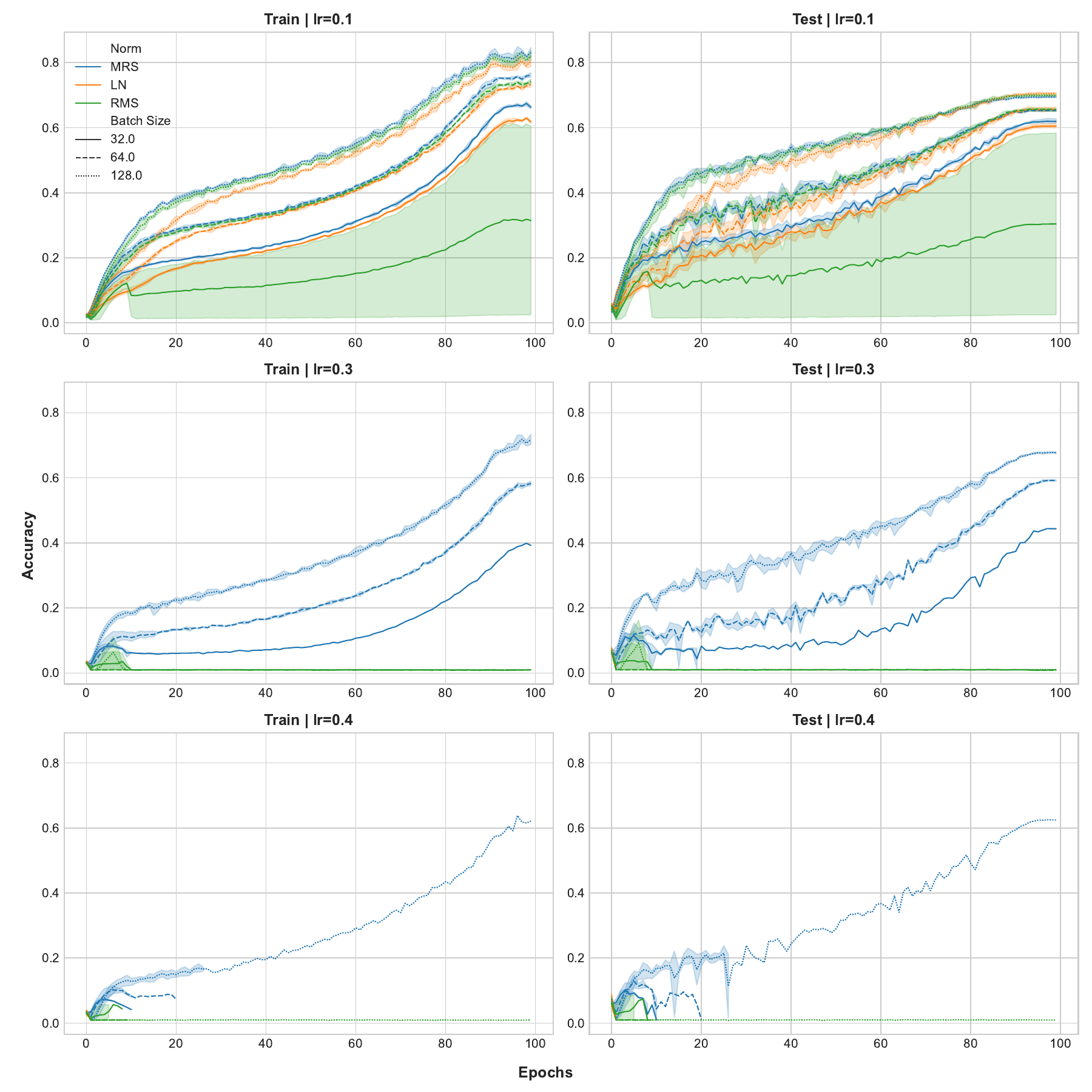}
    \caption{\textbf{Empirical evaluation of training stability under extreme hyperparameter regimes.}
    We compare MRSNorm (MRS, blue) against standard LayerNorm (LN, orange) and RMSNorm (RMS, green) across varying learning rates ($\text{lr} \in \{0.1, 0.3, 0.4\}$) and batch sizes ($\text{bs} \in \{128, 64, 32\}$).
    Line styles represent batch sizes (solid: $128$, dashed: $64$, dotted: $32$), and shaded regions denote variance.
    While LN and RMS suffer from catastrophic optimization collapse at higher learning rates ($\text{lr} \ge 0.3$), MRSNorm maintains a stable gradient flow, successfully converging for a batch size of $128$ at $\text{lr}=0.3$. Notably, under the most extreme setting ($\text{lr}=0.4$), MRSNorm is the only method that prevents immediate divergence.}
    \label{fig:train_curves}
\end{figure}

\paragraph{Robustness Under Extreme Optimization Regimes}
To empirically validate the theoretical stability of the phasor manifold introduced by MRSNorm, we designed a series of stress-test benchmarks.
The primary objective is to evaluate the resilience of normalization methods against severe optimization noise, exacerbated by high learning rates and small batch sizes.

\paragraph{Experimental Setup.}
We trained a ResNet model on the CIFAR-100 dataset \cite{krizhevsky2009learning}, varying the learning rate $lr \in \{0.1, 0.3, 0.4\}$ and batch size $bs \in \{128, 64, 32\}$.
We employed the SGD optimizer with a momentum of $0.9$ and weight decay of $5e^{-4}$.
We utilized 10 warmup epochs and applied cosine annealing for the learning rate schedule.
Each configuration was repeated across three random seeds (42, 43, 44) to ensure statistical significance, with the mean and variance of the training accuracy recorded.
We compared our proposed MRSNorm against two ubiquitous baseline normalizations: LayerNorm (LN) and RMSNorm (RMS).
All models were trained from scratch for 100 epochs.

\paragraph{Results in Standard Regimes ($lr=0.1$).}
As illustrated in the top panel of Figure \ref{fig:train_curves}, under a standard learning rate ($lr=0.1$), all normalization methods converge effectively at larger batch sizes ($bs \in \{64, 128\}$).
However, when the batch size is reduced to $bs=32$ (dotted lines)—thereby introducing significant stochastic noise into the gradient estimates—the structural vulnerability of 1D scalar normalization methods becomes apparent.
RMSNorm exhibits severe instability with high variance (indicated by the green shaded area) and fails to optimize effectively.
In contrast, MRSNorm maintains a smooth and stable training trajectory, outperforming both LN and RMS.

\paragraph{Catastrophic Collapse of Baselines in Extreme Regimes ($lr \ge 0.3$).}
The fundamental geometric superiority of MRSNorm is decisively demonstrated when optimization conditions are pushed to extreme limits.
As illustrated in Figure \ref{fig:train_curves} ($lr=0.3$), both LayerNorm and RMSNorm suffer from catastrophic optimization collapse.
Remarkably, MRSNorm remains robust under these severe hyperparameter regimes.
It successfully optimizes with a batch size of $128$, and even though the training curves exhibit oscillations at smaller batch sizes ($bs=64$ and $bs=32$), several seeds still manage to converge.
In contrast, the baseline normalization methods fail entirely, with their training accuracy immediately flattening near zero.

Even at the extreme learning rate of $lr=0.4$—four times the standard baseline setting—MRSNorm sustains initial gradient flow.
Although configurations with smaller batch sizes eventually diverge before 30 epochs, MRSNorm demonstrates significant endurance.
Conversely, the other normalization methods collapse instantaneously, failing to survive beyond the first few epochs.

\section{Discussion}
\subsection{The Paradox of Constraint: Analytical Contrast of RMSNorm and MRSNorm Gradients}
\label{sec:grad}
To mathematically unveil why MRSNorm sustains identical representational capacity and optimization stability despite having exactly halved affine parameters, we conduct a comparative analysis of the backpropagation dynamics of standard RMSNorm and our proposed MRSNorm. 

Under standard RMSNorm, which operates on independent 1D scalar channels $x_i \in \mathbb{R}$, the gradient of the loss $\mathcal{L}$ with respect to the input $x_i$ is formulated as:
\begin{equation}
    \label{eq:rms_dev}
    \frac{\partial \mathcal{L}}{\partial x_i} = \frac{1}{S_{RMS}} \left[ \nabla o_i \gamma_i - \mathbf{\hat{x}_i} \left( \frac{1}{N} \sum_{k=1}^N \gamma_k \nabla o_k \hat{x}_k \right) \right]
\end{equation}
where $N = 2P$ is the number of independent channels.
In contrast, under the $2D$ phasor-based MRSNorm, the gradient vector of the loss with respect to the input phasor $\mathbf{z}_i = [x_i, y_i]^T \in \mathbb{R}^2$ is expressed as:
\begin{equation}
    \label{eq:mrs_dev}
    \frac{\partial \mathcal{L}}{\partial \mathbf{z}_i} = \frac{1}{S_{MRS}} \left[ \nabla \mathbf{o}_i \gamma_i - \mathbf{u}_i \left( \frac{1}{P} \sum_{k=1}^P \gamma_k \big( \nabla \mathbf{o}_k \cdot \mathbf{\hat{z}}_k \big) \right) \right]
\end{equation}
where $\mathbf{u}_i = [\cos\theta_i, \sin\theta_i]^T$ is the unit direction vector of the $i$-th phasor, and $\cdot$ represents the standard Euclidean dot product in $\mathbb{R}^2$.

The comparative contrast between Equation \ref{eq:rms_dev} and Equation \ref{eq:mrs_dev} reveals three numerical-analytical insights:

First, we identify a \textbf{Global-Local Paradox} in gradient scale propagation. 
In standard RMSNorm, the scalar gradient update is scaled by the raw, normalized coordinate magnitude $\mathbf{\hat{x}_i}$. If $x_i$ acts as an unbounded outlier, its gradient scales proportionally, causing it to monopolize the backpropagated learning signal and starve the remaining channels.
Conversely, if $x_i$ falls near the origin, its gradient collapses, leading to local representation underflow. 

Let $\Omega = \frac{1}{P} \sum_{k=1}^P \gamma_k \big( \nabla \mathbf{o}_k \cdot \mathbf{\hat{z}}_k \big) \in \mathbb{R}$ denote the \textbf{global gradient projection scalar}, which represents the ensemble-averaged error projection onto the phasor manifold. 

In MRSNorm, the unstable and outlier-prone multiplier $\hat{x}_i$ of RMSNorm is replaced by the unit direction vector $\mathbf{u}_i$.
Because $\|\mathbf{u}_i\|_2 = \sqrt{\cos^2\theta_i + \sin^2\theta_i} \equiv 1$ unconditionally, the $L_2$ norm of the local gradient interference vector is mathematically guaranteed to be uniform and invariant to the phase angle $\theta_i$:
\begin{equation}
    \left\| \frac{\Omega}{S_{MRS}} \mathbf{u}_i \right\|_2 = \frac{|\Omega|}{S_{MRS}}
\end{equation}
This local isotropic normalization ensures that every single $2D$ phasor, regardless of its raw activation scale or angular orientation, receives an identical magnitude of gradient updates.
This \textbf{Gradient Homogeneity} proactively prevents both local gradient starvation and scale explosion.

Second, MRSNorm guarantees \textbf{Conformal Gradient Collinearity}. Because the subtraction term in Equation \ref{eq:mrs_dev} is multiplied directly by the unit vector $\mathbf{u}_i$, the scale adjustment is applied strictly parallel to the input phasor $\mathbf{z}_i$ itself:
\begin{equation}
    \label{eq:parallel_grad}
    \mathbf{u}_i \cdot \text{Scale} \parallel \mathbf{z}_i
\end{equation}
Subtracting a collinear vector from a $2D$ phasor strictly alters its radial magnitude (amplitude) while leaving its phase angle ($\theta_i$) mathematically decoupled.
Standard RMSNorm, by scaling coordinates independently ($\gamma_{x, i} \neq \gamma_{y, i}$), deforms the circular aspect ratio into an ellipse, corrupting the phase. 

Third, this geometric purity explains why MRSNorm matches the performance of RMSNorm despite having \textbf{halved affine parameters}.
The independent scaling of $x$ and $y$ in standard norms is not an asset, but a \textbf{harmful redundancy} that merely serves to inject phase-domain noise.
By eliminating this redundant parameterization and restricting the update to a conformal, phase-preserving manifold, MRSNorm simplifies the optimization landscape, allowing the model to converge into flatter, more robust minima with half the parameter footprint.

\subsection{Mathematical Foundations: Why $2D$ Phasor Pairing is a Structural Necessity}
\label{sec:math_foundation}

Our architecture relies on two mathematical pillars: why phase-stable normalization is impossible in $1D$, and why strict amplitude normalization fails via radial gradient blockade.

First, we prove that stable periodic gradient propagation is structurally impossible in $1D$ real analysis. 
Let $f: \mathbb{R} \to \mathbb{R}$ be a continuously differentiable, periodic activation function with a constant gradient norm $|f'(x)| = c > 0$.
By the Intermediate Value Theorem, a continuous derivative with a constant absolute value must be either $c$ or $-c$.
Integrating this yields a linear function ($f(x) = \pm cx + d$), which is inherently non-periodic.
Thus, no $1D$ scalar channel can simultaneously exhibit periodicity and gradient stability.
MRSNorm overcomes this bottleneck by ascending to the $2D$ phasor space $\mathbb{R}^2$.
By pairing channels, we exploit the Pythagorean identity $\|\mathbf{u}_i\|_2 = \sqrt{\cos^2\theta_i + \sin^2\theta_i} \equiv 1$, allowing individual components to behave as smooth, periodic functions while maintaining a strictly constant update norm.

Second, we address why we must avoid strict instance-wise projection onto the unit circle.
Let $\mathbf{\hat{z}}_i = \mathbf{z}_i / \|\mathbf{z}_i\|_2$.
The Jacobian $\mathbf{J}_i = \frac{\partial \mathbf{\hat{z}}_i}{\partial \mathbf{z}_i} = \frac{1}{\|\mathbf{z}_i\|_2} [\mathbf{I} - \mathbf{u}_i \mathbf{u}_i^T]$ contains a rank-1 projection matrix $\mathbf{P}_i = \mathbf{I} - \mathbf{u}_i \mathbf{u}_i^T$.
Since $\mathbf{P}_i \mathbf{u}_i = \mathbf{0}$, any backpropagating error signal along the radial direction is identically annihilated ($\nabla \mathbf{\hat{z}}_i^T \mathbf{J}_i \mathbf{u}_i = 0$).
This \textit{Radial Gradient Blockade} renders deep training impossible by severing the optimization path for amplitude parameters. 

MRSNorm resolves this via ensemble-based scaling.
By normalizing with the global $L_1$ average magnitude $S_{MRS}$ rather than the instance-specific $\|\mathbf{z}_i\|_2$, the Jacobian takes the form:
\begin{equation}
    \frac{\partial \mathbf{\hat{z}}_i}{\partial \mathbf{z}_i} = \frac{1}{S_{MRS}} \left[ \mathbf{I} - \frac{1}{P} \mathbf{u}_i \mathbf{\hat{z}}_i^T \right]
\end{equation}
The projection term is scaled by $1/P$, leaving the radial gradient unblocked and stable as $P$ increases.
Consequently, $2D$ phasor pairing is not merely a design choice, but the minimal viable dimension to preserve both phase periodicity and radial gradient flow, providing the theoretically sound foundation.

\subsection{Revisiting Attention Mechanisms: Geometric Resolution of Logit Imbalance via MRSNorm}
\label{sec:revisiting_attention}
Standard scaled dot-product attention fundamentally relies on the dot product between queries ($\mathbf{q}$) and keys ($\mathbf{k}$).
When specific tokens disproportionately expand their magnitudes, they dominate the attention logits regardless of semantic direction, driving the softmax function into a near one-hot state (softmax collapse) and causing severe gradient vanishing.

By integrating an affine-free MRSNorm immediately after the linear projections (QK-MRSNorm), we introduce a \textit{soft geometric constraint} that alleviates this issue.
MRSNorm binds the pairwise channels into a phasor manifold.
To understand its geometric implication, consider a query vector decomposed into $d/2$ phasor blocks, where each 2D block $\mathbf{q}_i$ has a local magnitude $r_{q,i} = \|\mathbf{q}_i\|_2$ and a phase.
MRSNorm normalizes these blocks by the global Mean Root Square energy $S_q$.
Thus, the attention score between a query and a key evaluates to:

\begin{equation}
\text{Score} = \sum_{i=1}^{d/2} \hat{\mathbf{q}}_i \cdot \hat{\mathbf{k}}_i = \sum_{i=1}^{d/2} \left( \frac{r_{q,i} \cdot r_{k,i}}{S_q \cdot S_k} \right) \cos(\Delta\theta_i)
\end{equation}

This mathematical equivalence reveals an architectural property: QK-MRSNorm implicitly transforms the standard dot product into an \textbf{Energy-Weighted Cosine Similarity Operator}. 
Specifically, the term $\cos(\Delta\theta_i)$ captures the pure semantic phase alignment, while the magnitude ratio $(r_{q,i} r_{k,i}) / (S_q S_k)$ acts as an auto-gating mechanism.
The network learns to dynamically allocate magnitude to highly informative phasor blocks, amplifying their vote in the final score while suppressing irrelevant features.

Furthermore, the algebraic derivation demonstrating how MRSNorm disentangles the standard dot product into an energy-weighted cosine similarity is provided in Appendix \ref{app:qk_mrsnorm}. 
A formal mathematical proof establishing the strict numerical upper bound of this operator—and detailing its localized geometric superiority over QK-RMSNorm via the semantic phase veto—is presented in Appendix \ref{app:qk_stability}.
Together, these appendices theoretically validate both the enhanced representational capacity and the guaranteed numerical stability of the proposed QK-MRSNorm framework.

\section{Conclusion}
While this study conclusively demonstrates the theoretical and empirical robustness of MRSNorm, our evaluation primarily focuses on rigorous stress-testing of optimization dynamics under extreme hyperparameters (e.g., high learning rates and small batch sizes).
These micro-benchmarks are essential for proving the fundamental stability of the phasor manifold and the preservation of Jacobian isometry.
However, scaling this architectural modification to billion-parameter LLMs inherently requires substantial computational clusters.
Consequently, an empirical analysis of MRSNorm's impact on pre-training dynamics at a massive scale remains outside the scope of this work.

The introduction of MRSNorm opens promising avenues for future research, most notably in redesigning attention routing mechanisms.
As discussed in Section \ref{sec:revisiting_attention}, applying an affine-free MRSNorm directly to the queries and keys (QK-MRSNorm) mathematically transforms the standard scaled dot-product into an \textit{Energy-Weighted Cosine Similarity Operator}.
We theoretically demonstrated that this formulation intrinsically resolves logit imbalance and acts as an auto-gating mechanism via localized magnitude variations ($r_{q,i} r_{k,i}$). 
We strongly invite the community to empirically explore this QK-MRSNorm architecture in large-scale Transformer variants and state-space models.
Validating whether this geometric soft constraint can prevent softmax collapse in long-context attention will be a critical next step.
We believe the geometric stability provided by MRSNorm will serve as a foundational building block for the next generation of robust neural architectures.

\bibliographystyle{unsrt}
\bibliography{ref}

\newpage
\appendix
\section{Appendix: Mathematical Derivation of Energy-Weighted Cosine Similarity via MRSNorm}
\label{app:qk_mrsnorm}

In Section \ref{sec:revisiting_attention}, we proposed that applying an affine-free MRSNorm to queries and keys (QK-MRSNorm) inherently transforms the standard dot-product attention into an \textit{energy-weighted cosine similarity} operator. This appendix provides the rigorous step-by-step mathematical derivation of this equivalence and details its optimization advantages over strict $L_2$ normalization.

\subsection{Step 1: Phasor Decomposition and Polar Representation}
Let $\mathbf{q}, \mathbf{k} \in \mathbb{R}^d$ be a query and a key vector, respectively, where the hidden dimension $d$ is an even number. MRSNorm processes channels in pairwise blocks. We can decompose the query vector into $d/2$ two-dimensional blocks, $\mathbf{q}_i \in \mathbb{R}^2$ for $i \in \{1, 2, \dots, d/2\}$.

Each 2D block can be uniquely represented in polar coordinates. Let $r_{q,i} = \|\mathbf{q}_i\|_2$ be the local magnitude (energy) of the block, and $\theta_{q,i}$ be its phase angle. Thus,
\begin{equation}
\mathbf{q}_i = \begin{bmatrix} r_{q,i} \cos\theta_{q,i} \\ r_{q,i} \sin\theta_{q,i} \end{bmatrix}, \quad \mathbf{k}_i = \begin{bmatrix} r_{k,i} \cos\theta_{k,i} \\ r_{k,i} \sin\theta_{k,i} \end{bmatrix}
\end{equation}

\subsection{Step 2: Global RMS Normalization (Soft Constraint)}
MRSNorm calculates the global Root Mean Square (RMS) energy across all blocks. For the query vector, the scaling factor $S_q$ is defined as:
\begin{equation}
S_q = \sqrt{\frac{1}{d/2} \sum_{i=1}^{d/2} \|\mathbf{q}_i\|_2^2 + \epsilon} = \sqrt{\frac{1}{d/2} \sum_{i=1}^{d/2} r_{q,i}^2 + \epsilon}
\end{equation}
The normalized vector $\hat{\mathbf{q}}$ consists of blocks $\hat{\mathbf{q}}_i = \frac{\mathbf{q}_i}{S_q}$. Notice that each block is scaled by the \textit{global} energy $S_q$, not its \textit{local} energy $r_{q,i}$. This is the fundamental distinction between MRSNorm's soft constraint and strict $L_2$ projection.

\subsection{Step 3: Dot Product Expansion and Equivalence}
The attention score before applying softmax is the dot product of the normalized vectors $\hat{\mathbf{q}}$ and $\hat{\mathbf{k}}$:
\begin{equation}
\label{eq:qk_mrsnorm_dot}
\text{Score} = \hat{\mathbf{q}} \cdot \hat{\mathbf{k}} = \sum_{i=1}^{d/2} \hat{\mathbf{q}}_i \cdot \hat{\mathbf{k}}_i = \sum_{i=1}^{d/2} \left( \frac{\mathbf{q}_i}{S_q} \right) \cdot \left( \frac{\mathbf{k}_i}{S_k} \right) = \frac{1}{S_q S_k} \sum_{i=1}^{d/2} \mathbf{q}_i \cdot \mathbf{k}_i
\end{equation}
Expanding the dot product of the individual 2D blocks using their polar representations:
\begin{align}
\label{eq:qk_mrsnorm_dot_product_expansion}
\mathbf{q}_i \cdot \mathbf{k}_i &= (r_{q,i} \cos\theta_{q,i})(r_{k,i} \cos\theta_{k,i}) + (r_{q,i} \sin\theta_{q,i})(r_{k,i} \sin\theta_{k,i}) \nonumber \\
&= r_{q,i} r_{k,i} \left( \cos\theta_{q,i} \cos\theta_{k,i} + \sin\theta_{q,i} \sin\theta_{k,i} \right) \nonumber \\
&= r_{q,i} r_{k,i} \cos(\theta_{q,i} - \theta_{k,i})
\end{align}
Let $\Delta\theta_i = \theta_{q,i} - \theta_{k,i}$ be the phase difference (semantic alignment) between the query and key for the $i$-th micro-head. Substituting Eq. \ref{eq:qk_mrsnorm_dot_product_expansion} back into Eq. \ref{eq:qk_mrsnorm_dot}, we arrive at the exact formulation presented in the main text:
\begin{equation}
\label{eq:qk_mrsnorm_score}
\text{Score} = \sum_{i=1}^{d/2} \underbrace{\left( \frac{r_{q,i}}{S_q} \frac{r_{k,i}}{S_k} \right)}_{\text{Energy Weight } W_i} \cdot \underbrace{\cos(\Delta\theta_i)}_{\text{Phase Alignment}}
\end{equation}

\subsection{Step 4: Computational Equivalence and Zero-Overhead Implementation}
It is crucial to emphasize that Eq. \ref{eq:qk_mrsnorm_score} represents a mathematical interpretation of the geometric manifold, \textbf{not a required computational graph}. In practice, the forward pass never explicitly computes polar coordinates, local magnitudes ($r_{q,i}$), or trigonometric functions ($\cos\Delta\theta_i$).

Because the formulations are strictly mathematically equivalent, one only needs to apply the standard linear dot product operation ($\hat{\mathbf{q}} \cdot \hat{\mathbf{k}}$) over the MRSNorm-processed vectors. The geometric constraints embedded in the MRSNorm manifold naturally guarantee that this simple inner product behaves exactly as the complex, auto-gated ensemble of micro-heads derived in Eq. \ref{eq:qk_mrsnorm_score}. Consequently, QK-MRSNorm endows the attention mechanism with robust representation power and dynamic confidence routing with \textbf{zero additional computational overhead or latency}.

\section{Mathematical Proof of Logit Stability in QK-MRSNorm}
\label{app:qk_stability}

In Section \ref{sec:revisiting_attention}, we discussed how QK-MRSNorm geometrically resolves logit imbalance. Here, we formally prove its strict upper bound and demonstrate its theoretical superiority over RMSNorm, highlighting the critical role of the continuous phase in neutralizing magnitude spikes.

\paragraph{Limitations of QK-RMSNorm: Unilateral Magnitude Dominance.}
Recently, QK-RMSNorm has been adopted to mitigate logit explosion by applying 1D scalar normalization.
However, RMSNorm is fundamentally limited by its lack of geometric dimensionality.
Because a 1D scalar possesses only a binary sign ($\pm 1$), the interaction between a query and key ($\hat{q}_i \hat{k}_i$) is entirely dictated by their normalized magnitudes.
If a single channel monopolizes the overall variance and spikes, this magnitude surge unilaterally forces a massive localized inner product.
Without a mechanism to decouple magnitude from semantic direction, an anomalously large scalar spike bypasses all semantic filters, inevitably monopolizing the global attention score.

\paragraph{Local Stability via Semantic Phase Veto.}
In contrast, QK-MRSNorm binds pairwise channels into a continuous phasor manifold, transforming how local features interact.
The local inner product $P_i$ between a single query and key phasor block evaluates to:
\begin{equation}
P_i = \hat{\mathbf{q}}_i \cdot \hat{\mathbf{k}}_i = \left( \frac{r_{q,i} \cdot r_{k,i}}{S_q \cdot S_k} \right) \cos(\Delta\theta_i)
\end{equation}
This geometric decoupling introduces a profound architectural safeguard: the \textit{Semantic Phase Veto}.
Unlike 1D scalars that blindly multiply their magnitudes, the magnitude term in a phasor is merely a potential energy.
Even if the magnitude weight $(r_{q,i} r_{k,i}) / (S_q S_k)$ reaches its structural extreme, this surge is completely neutralized if the semantic directions are orthogonal ($\cos(\Delta\theta_i) \to 0$).
In other words, a magnitude spike cannot brute-force its way into the attention score; it is strictly gated by the continuous semantic alignment of the features.
The localized contribution is therefore bounded not just structurally, but semantically:
\begin{equation}
|P_i| \le \frac{r_{q,i}}{S_q} \frac{r_{k,i}}{S_k} \le \frac{d}{2}
\end{equation}

\paragraph{Global Bounding via Cauchy-Schwarz Inequality.}
Guarded by this semantic phase veto at the micro-level, the global attention score—the sum of all localized phasor interactions—is mathematically guaranteed to remain exceptionally stable.
Applying the Cauchy-Schwarz inequality to the sum of the magnitudes yields:
\begin{equation}
|\text{Score}| = \left| \sum_{i=1}^{d/2} P_i \right| \le \sum_{i=1}^{d/2} \frac{r_{q,i} \cdot r_{k,i}}{S_q \cdot S_k} \le \frac{1}{S_q \cdot S_k} \sqrt{\sum_{i=1}^{d/2} r_{q,i}^2} \sqrt{\sum_{i=1}^{d/2} r_{k,i}^2}
\end{equation}
By the definition of the global RMS energy, $\sum r_{q,i}^2 = \frac{d}{2}S_q^2$ and $\sum r_{k,i}^2 = \frac{d}{2}S_k^2$.
Substituting these terms simplifies the absolute upper bound to:
\begin{equation}
|\text{Score}| \le \frac{1}{S_q \cdot S_k} \left( \sqrt{\frac{d}{2}} S_q \cdot \sqrt{\frac{d}{2}} S_k \right) = \frac{d}{2}
\end{equation}

This rigorous derivation demonstrates that QK-MRSNorm establishes absolute geometric stability.
While the Cauchy-Schwarz inequality guarantees a structural maximum logit of $d/2$, the true architectural breakthrough lies in the phase veto.
Regardless of how extremely an individual feature expands, it cannot hijack the attention mechanism without strict semantic alignment.
This dual protection—semantic gating at the local level and absolute bounding at the global level—structurally alleviates the possibility of magnitude-driven softmax collapse.

\section{Implementation of GroupMRSNorm}
\label{sec:imp}
\begin{lstlisting}[style=mycode]
import torch
import torch.nn as nn
from typing import Union, Tuple, List
import math


class GroupMRSNorm(nn.Module):

    def __init__(
        self, num_groups: int, num_channels: int, channel_dim: int, eps: float = 1e-6
    ):
        super().__init__()
        if num_channels % 2 != 0:
            raise ValueError(
                f"channel ({num_channels}) must be even for 2D Phasor pairs."
            )

        num_bundles = num_channels // 2
        if num_bundles % num_groups != 0:
            raise ValueError(
                f"number of hidden bundles ({num_bundles}) must be divisible by the specified number of groups ({num_groups})."
            )

        self.num_groups = num_groups
        self.num_channels = num_channels
        self.num_bundles = num_bundles
        self.channel_dim = channel_dim
        self.eps = eps

        self.weight = nn.Parameter(torch.empty(num_bundles))
        self.reset_parameters()

    def reset_parameters(self) -> None:
        nn.init.ones_(self.weight)

    def forward(self, x: torch.Tensor) -> torch.Tensor:
        orig_dtype = x.dtype
        orig_shape = x.shape

        x_fp32 = x.to(torch.float32)

        c_dim = (
            self.channel_dim
            if self.channel_dim >= 0
            else x_fp32.dim() + self.channel_dim
        )
        p_dim = c_dim + 1

        paired_shape = (
            list(orig_shape)[:c_dim]
            + [self.num_bundles, 2]
            + list(orig_shape)[c_dim + 1 :]
        )
        x_paired = x_fp32.view(*paired_shape)

        mag = torch.norm(x_paired, p=2, dim=p_dim, keepdim=True)

        C_per_G = self.num_bundles // self.num_groups
        grouped_mag_shape = (
            list(mag.shape)[:c_dim]
            + [self.num_groups, C_per_G]
            + list(mag.shape)[c_dim + 1 :]
        )
        mag_g = mag.view(*grouped_mag_shape)

        reduce_dims = tuple(range(c_dim + 1, mag_g.dim()))
        l1_mean_g = torch.mean(mag_g, dim=reduce_dims, keepdim=True) + self.eps

        x_grouped_shape = (
            list(x_paired.shape)[:c_dim]
            + [self.num_groups, C_per_G]
            + list(x_paired.shape)[c_dim + 1 :]
        )
        x_g = x_paired.view(*x_grouped_shape)
        x_normed_g = x_g / l1_mean_g

        x_normed = x_normed_g.view(*orig_shape)

        weight_expanded = torch.repeat_interleave(self.weight, repeats=2, dim=0)
        view_shape = [1] * x.dim()
        view_shape[c_dim] = self.num_channels
        weight_view = weight_expanded.view(*view_shape)

        out = x_normed * weight_view

        return out.to(orig_dtype)
\end{lstlisting}

\end{document}